\documentclass[journal]{IEEEtran}
\usepackage{cite}
\usepackage{amsmath}
\usepackage{graphicx}
\usepackage{wrapfig}
\usepackage{array}
\usepackage{multirow,multicol}
\usepackage{pifont}
\usepackage{float}
\usepackage{hyperref}
\usepackage{xcolor}
\hypersetup{
    colorlinks,
    linkcolor={red!50!black},
    citecolor={blue!50!black},
    urlcolor={blue!80!black}
}
\usepackage{bm}

\usepackage{dblfloatfix}

\begin{document}

\title{Locally Adaptive Decay Surfaces for High-Speed Face and Landmark Detection with Event Cameras}
\author{Paul Kielty*,Timothy Hanley, \IEEEmembership{Graduate Student Member, IEEE},
Peter Corcoran, \IEEEmembership{Fellow, IEEE}%
\thanks{*Corresponding author. Email: paul.p.kielty@universityofgalway.ie.}%
\thanks{With the College of Science and Engineering, University of Galway, Ireland.}}

\maketitle

\begin{abstract}
Event cameras record luminance changes with microsecond resolution, but converting their sparse, asynchronous output into dense tensors that neural networks can exploit remains a core challenge. Conventional histograms or globally-decayed time-surface representations apply fixed temporal parameters across the entire image plane, which in practice creates a trade-off between preserving spatial structure during still periods and retaining sharp edges during rapid motion. We introduce Locally Adaptive Decay Surfaces (LADS), a family of event representations in which the temporal decay at each location is modulated according to local signal dynamics. Three strategies are explored, based on event rate, Laplacian-of-Gaussian response, and high-frequency spectral energy. These adaptive schemes preserve detail in quiescent regions while reducing blur in regions of dense activity. Extensive experiments on the public data show that LADS consistently improves both face detection and facial landmark accuracy compared to standard non-adaptive representations. At 30~Hz, LADS achieves higher detection accuracy and lower landmark error than either baseline, and at 240~Hz it mitigates the accuracy decline typically observed at higher frequencies, sustaining 2.44\% normalized mean error for landmarks and 0.966~mAP\textsubscript{50} in face detection. These high-frequency results even surpass the accuracy reported in prior works operating at 30~Hz, setting new benchmarks for event-based face analysis. Moreover, by preserving spatial structure at the representation stage, LADS supports the use of much lighter network architectures while still retaining real-time performance. These results highlight the importance of context-aware temporal integration for neuromorphic vision and point toward real-time, high-frequency human-computer interaction systems that exploit the unique advantages of event cameras.
\end{abstract}




\section{Introduction}
\label{sec:introduction}
Event cameras sense changes in pixel intensity asynchronously, producing sparse streams of temporally precise events rather than dense image frames. This sensing paradigm offers significant advantages for computer vision, including high temporal resolution, low latency, wide dynamic range, and efficient data capture. These properties make event cameras particularly attractive for embedded and real-time applications in robotics, driver monitoring, and human-computer interaction \cite{gallego2020event, shariff2024Automotive, EX-Gaze}.

Despite these benefits, neural networks in conventional computer vision expect dense, spatially structured inputs. To take advantage of the advancements in this more mature field, the sparse event streams are often condensed into 2D or 3D arrays that capture spatiotemporal structure. Such transformations fall under the broad category of event "representations" \cite{gallego2020event}. Popular approaches include event histograms, which summarize activity over time by accumulating events, and leaky time-surfaces which implement a decaying factor to past events. However, these methods typically apply fixed integration windows and uniform temporal decay across the entire image plane. This global treatment fails to account for the highly localized and dynamic nature of event generation, especially in tasks such as facial analysis where informative motion can have large variances in rate and may be confined to small regions of the scene. These are some of the reasons that the construction of effective intermediate representations remains a persistent challenge in event-based vision.

To address these limitations, we propose a new representation framework, Locally Adaptive Decay Surfaces (LADS), in which the temporal decay rate at each pixel is modulated according to local signal dynamics. LADS enables slower decay in low-activity regions to preserve weak but potentially meaningful signals, while applying faster decay in high-activity regions where dense event clusters can otherwise lead to blurring and accumulated noise. This mechanism ensures temporal precision in active areas and maintains spatial consistency in static regions, improving the overall informativeness of the event representation. For example, in an event video where the face remains mostly stationary apart from a blink, the regions around the eyes exhibit high signal dynamics due to the rapid motion of the eyelids. In these regions, slower decay would cause features from different phases of the blink to accumulate and blur together, making it difficult to localize landmarks such as the eyelid contours or pupil. Faster decay prevents this by emphasizing the most recent motion, while slower decay in the static parts of the face helps preserve consistent structure in regions like the nose or jawline.

The effectiveness of LADS is evaluated on two tasks, first the detection of facial bounding boxes, then the detection of facial landmarks within face crops. These two tasks together allow us to assess how adaptive decay influences both fine-grained localization and object-level detection performance.

Experiments are conducted on the publicly available Faces in Event Streams (FES) dataset. Noting the presence of numerous flawed samples in previous work \cite{kielty_lm+blink}, we published an exclusion list based on manual inspection. For this study, we extended our efforts by first developing an automated filtering algorithm to identify problematic samples, then conducting a targeted manual review to catch additional flawed examples not flagged by the algorithm. The final validated sample list is published alongside this paper.

The primary contributions of this paper are as follows:
\begin{itemize}
    \item We introduce LADS, a family of spatially adaptive event representations, and compare three specific measurement functions for local signal dynamics.
    \item We evaluate LADS on face detection and facial landmark localization, showing consistent gains over two common event representations.
    \item We demonstrate that LADS preserves accuracy at high update rates, better exploiting the event camera's temporal resolution while supporting the use of lighter network architectures.
\end{itemize}

The remainder of this paper is organized as follows: Section II reviews related work on event representations and the detection of faces and facial landmarks in events. Section III introduces our LADS framework along with an overview of our data, the network model, and the experimental procedures for benchmarking our method. Section V presents our results and analysis, and Section VI concludes the paper with a summary and discussion of future directions. The LADS toolkit is released publicly at \href{https://github.com/C3Imaging/LADS}{\textcolor{blue}{\underline{https://github.com/C3Imaging/LADS}}}.
\section{Related Work}
In this section, we review prior research on event-based representations, temporal integration strategies, and facial landmark detection in neuromorphic vision systems. We highlight the limitations of existing methods and situate our approach within the broader context of event-based computer vision.

\subsection{Event Representations}
Each event is defined as $e_{n} = (x_{n},y_{n},t_{n},p_{n})$, where $x_{n},y_{n}$ are the spatial coordinates of the event, $t_{n}$ is the event's timestamp, and  $p_{n} \in {-1, +1}$ is the event's polarity. Event‐based vision systems typically convert the sparse, asynchronous stream of events into dense, frame‐like tensors. One of the earliest approaches is the event image or frame, which marks whether a pixel fired at least once within a given window, defined either by a fixed duration or a fixed event count \cite{shariff2024Automotive}. Implementations may either ignore polarity and record binary activity, or store positive and negative events in separate channels. This binary presence signal is easy to process with conventional pipelines, though the magnitude of activity is not fully represented.

Building on this idea is the event histogram, a very popular representation that also operates on fixed-duration or fixed-count windows but, instead of simply marking activity, records the number of events per pixel in a more expressive form. Depending on the variant, this can be implemented by ignoring polarity \cite{liu2018_hist_polarity}, by keeping separate channels for positive and negative events \cite{ryan2021real}, or by summing the polarities into a single channel that reflects the cumulative polarity change over the window \cite{Kielty2023}. The latter is the form used in this work. Let $\mathcal{E}_k(x, y)$ denote the set of all events at pixel $(x, y)$ within window $k$. The value for each pixel in the histogram of this window is computed by

\begin{equation} \label{eq:hist}
H_k(x, y) = \sum_{e_n \in \mathcal{E}_{k}(x,y)} p_n,
\end{equation}

which captures both the magnitude and sign of recent changes in the scene to a single channel. However, because it discards events outside the current window, the histogram encodes no recency beyond the window boundaries and can lack spatial structure when motion is weak or intermittent.

An influential alternative is the time surface, introduced in the HOTS model \cite{HOTS}, which maintains a per-pixel memory by exponentially decaying the contribution of past events according to their time of arrival. This yields a smooth, recency-weighted map of activity that has proven effective for event-based feature extraction. HATS \cite{HATS} extended this concept by aggregating locally averaged time-surface values into spatial histograms, improving robustness to noise and timing jitter through pooling, and delivering stronger recognition performance on benchmark datasets.

Another interpretation of the time surface uses leaky integration (LI) of histograms \cite{Ryan2023}, which combines the per-window histogram with an exponentially decayed version of the previous integrated state:

\begin{equation} \label{eq:surface}
S_k(x, y) = H_k(x, y) + d_k \cdot S_{k-1}(x, y)
\end{equation}

Where $d_k$, the decay factor applied to the contribution from window $k-1$ for the creation of $S_k$, is given by

\begin{equation} \label{eq:decay_factor}
d_k =  \exp\left( -\frac{\Delta t_k}{\tau}\right)
\end{equation}

where $\Delta t_k$ is the elapsed time between the final event in window $k-1$ and the final event in window $k$, and $\tau$ is a decay time constant used to control the rate of decay. This retains the exponential memory of classical time surfaces but applies it to windowed histograms rather than to per-event timestamps. It is computationally efficient, as the update is performed once per window, and it adds weight to pixels with high activity rather than recency alone. Prior event-based facial analysis works have made effective use of this leaky form \cite{Ryan2023}, including our work with the FES dataset \cite{kielty_lm+blink} where LI outperformed the standard histogram for landmark detection, but the histogram proved more effective for precise detection of rapid blink sequences. The histogram can be regarded as a limiting case of LI in which no contribution from the previous state is retained. Given these complementary strengths, and the fact that the histogram is already a prerequisite in our proposed methods, both are included as baselines for the present study. For clarity in the results and discussion, this time surface method is referred to as global-LI to distinguish it from the locally adaptive variants introduced in Section~\ref{sec:methods}.

Wang et al. introduce the Adaptive Gradient‐Based Time Surface \cite{adaptive_gradient_timesurfs}, which computes spatial–temporal gradients on the surface of active events to estimate local motion speed and then uses those gradient magnitudes to rescale the decay constant at each pixel. By doing so, they preserve fine texture in slowly moving regions while preventing motion blur in fast‐moving areas, yielding a 1.52\% improvement on the Gen1 automotive dataset and a 7.7\% gain on the RotateDigit benchmark compared to uniform‐decay surfaces. These findings highlight the value of motion‐aware temporal integration, though it was introduced late in our experiments and is not included for direct comparison.

\begin{figure*}[!b]
    \centering
    \includegraphics[width=1\linewidth]{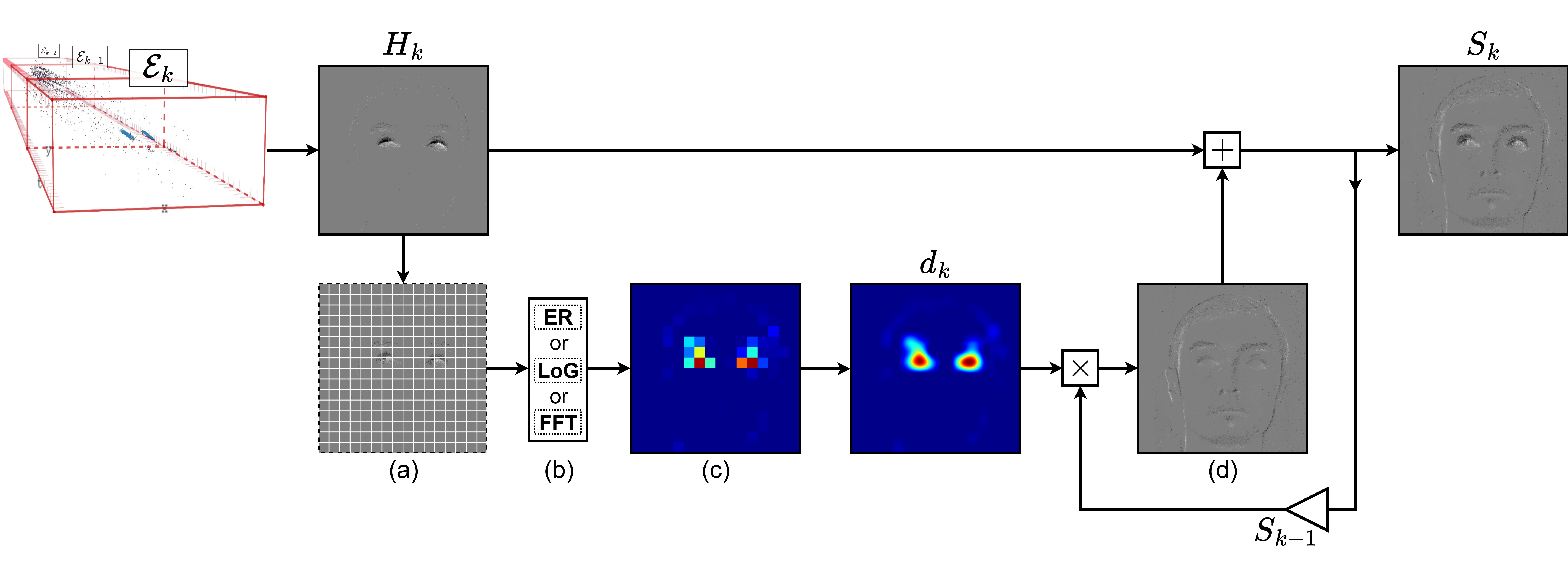}
    \caption{Visualization of the LADS process described in Section \ref{sec:measures}. The input window of events, $\boldsymbol{\mathcal{E}}_k$, is accumulated into histogram $H_k$. The subsequent annotated steps are: (a) divide the histogram into a user-specified number of patches; (b) measure the signal dynamics of patch according to the selected LADS variant; (c) compute a decay value for each patch, which are interpolated to generate the per-pixel decay map $d_k$; (d) decay the previous surface, $S_{k-1}$, by multiplication with $d_k$ (element-wise). The resulting product is added to $H_k$ to construct the new surface, $S_k$.}
    \label{fig:LADS_process}
\end{figure*}

\subsection{Event-Based Face Detection and Landmark Localization}
Face detection represents a crucial computer vision task with a wide range of human-focused applications. It is often required for further downstream tasks such as facial analytics, classification, or landmark detection. However, the asynchronous nature of event cameras makes it a challenging task, due to the face not necessarily being persistent in the event stream. Therefore, various representation approaches have been applied by different works in this area. 

Initial work in this area by \cite{face1} proposed a random forest detector, applied to events accumulated in a 60~ms window based on a histogram of oriented gradients \cite{hog}. This work showed the feasibility of face detection from event streams, demonstrating good performance on a small dataset. \cite{face2} proposed a face detection and tracking approach that detected the unique temporal signature of blinks in event streams to detect blinks and then subsequently track the face. 

More recently, \cite{ryan2021real} proposed a gated recurrent YOLO network along with a synthetic event dataset, Neuromorphic-HELEN for face and blink detection in events. The model, applied to a fixed count (50,000) event histogram, demonstrated strong results when tested on real event data containing both slight and larger head movements. However, the authors noted the lack of a large-scale publicly available annotated event dataset as a hindrance to work in this field.

Event cameras have also become increasingly attractive for a range of facial analytics tasks due to their high temporal resolution, low latency, and robustness to both challenging illumination and fast motion. These properties make event-based sensors particularly well-suited for real-time analysis of facial activity, including eye-blink detection, expression analysis, and gaze or pupil tracking, especially in situations where conventional frame-based cameras struggle with motion blur or dynamic lighting. Despite this promise, the application of event-based cameras to precise facial landmark localization remains relatively underexplored. Only a limited number of studies have directly addressed the problem of detecting and tracking facial landmarks from event streams, in part due to the technical challenges associated with sparse, asynchronous data and the lack of public benchmark datasets \cite{berlincioni2023neuromorphic}. 

An early event work that incorporates facial landmarks is NeuroDFD by Liu et al. \cite{liu2022neurodfd}. Their method constructs event volumes from fixed-count windows of 15,000 events with five temporal bins, handling positive and negative polarities separately before concatenation and percentile normalization. To overcome the lack of annotated events, the authors generated pseudo-labels for faces and landmarks by aligning event volumes with conventional image frames. Evaluated on the NeuroIV dataset, NeuroDFD achieved real-time inference with a model sized just over 1~MB , outperforming frame-based baselines. However, the authors note that landmark accuracy was constrained by the event accumulation strategy, since each event window captures the change in landmark position over an interval rather than a precise point in time.

The advancement of deep learning methods for event-based facial analysis has been hampered by the lack of large, annotated datasets suitable for direct training and evaluation. The FES dataset \cite{Bissarinova2024} represents a substantial step forward, introducing the first large-scale, open-source benchmark specifically for face and landmark detection on event camera data. FES provides nearly 700 minutes of event camera recordings, in which over 1.6 million faces from 73 participants are annotated with bounding boxes and five facial landmarks.

In their accompanying study, the FES authors trained several deep learning models for face detection and landmark detection. Their approach uses event histograms as the primary input representation, aggregating events over fixed-duration windows (either 33, 50, or 100 ms). The authors employed a deep convolutional network based on \cite{eventobjectdetection} for combined face bounding box and landmark detection. Models were evaluated across both controlled and unconstrained settings, with face detection and landmark performance reported in terms of normalized mean error (NME). These baseline NME results, along with the open release of both data and trained models, provide a valuable reference point for subsequent research in event-based facial landmark detection.

Making use of FES for face detection, our recent article \cite{spiking_face} explores a hybrid spiking neural network and artificial neural network architecture to combine the respective efficiency and feature extraction capabilities of each. This work shows promise in deploying event-based face detection in power or computationally limited applications on the edge.
In a parallel work we extended event-based facial landmark detection to a full set of 98 landmarks by leveraging both real and synthetic event streams \cite{kielty_lm+blink}. We introduced a multi-task deep network trained for simultaneous landmark localization and blink detection, and demonstrated state-of-the-art performance on the FES benchmark. Our findings emphasized the importance of temporal integration, showing that leaky time-surface representations consistently improved landmark accuracy compared to fixed-window histograms.


\section{Methodology}
\label{sec:methods}
We start this section with an explanation of the overarching principle behind our localized decay procedure and discuss several proposed functions for measuring the signal dynamics. We follow with a description of the dataset used for evaluating our methods and an account of the cleaning and preprocessing it required. Finally, we outline the face and facial landmark detection network architectures used in our experiments with details of the training and evaluation protocols.

\subsection{Measurement of Local Signal Dynamics}
\label{sec:measures}
The core of our approach is to replace uniform temporal integration with a locally adaptive scheme to preserve informative regions until the next batch of relevant events arrives. For this, the event stream is divided into a grid of non-overlapping spatial patches. Within each patch, we compute a summary measure of signal dynamics to determine the appropriate decay rate for that region. The use of non-overlapping patches provides a balance between computational efficiency and spatial specificity. With just a single score per patch, we also get just one modified decay value per patch. However, assigning a single decay value uniformly to all pixels within each patch can introduce visible discontinuities at the borders between adjacent patches. These abrupt changes create artificial edges and can degrade the fidelity of the event representation. To address this, we apply bilinear interpolation to the grid of patch-wise decay values, generating a smooth, per-pixel decay field that is the same size as the original event image. This ensures that transitions in decay rates are gradual across the image. The step-by-step process is visualized in Fig. \ref{fig:LADS_process}.

The next step is to define an effective statistic for quantifying signal dynamics within each patch. In this work, we explore several approaches, each capturing a different aspect of local activity, to guide the adaptive decay process. All methods were designed to be scalable to variable patch sizes with minimal additional parameter tuning.

\subsubsection{Adaptive Decay via Event Rate}
The concept was first prototyped using the event rate (ER) within each patch, as a measure of activity to scale the decay factor $d$ of a standard leaky time-surface. In regions with high ERs, new information arrives rapidly, and the decay rate must be increased to avoid obscuring recent features with stale ones. Conversely, a low ER indicates little change, so the decay is slowed to prevent useful spatial information from fading prematurely. The ERs are calculated as events per pixel per second, allowing dynamic adjustment to various patch sizes and update frequencies. The method can be tuned by adjusting the baseline value of $ \tau $ as in (\ref{eq:surface}). However, since ERs can vary so significantly between datasets with different sensor bias settings and environmental conditions, a reference ER parameter, $\lambda_0$, was included for easier normalization and tuning. The histogram of new events, $H_k$, is calculated as before and then divided into non-overlapping patches. For each patch, $P$, we calculate the patch ER, $\lambda_P$, and define the decay value for the patch as:
\begin{equation}\label{eq:decay_event_rate}
d^{\mathrm{ER}}_P = \exp\left( -\frac{\Delta t}{\tau} \cdot \frac{\lambda_0}{\lambda_P} \right).
\end{equation}

The patchwise decay values $d^{\mathrm{ER}}_P$ are bilinearly interpolated to the original event image resolution, yielding a per-pixel decay field $d_k^{\mathrm{ER}}(x, y)$ for each window $k$. The resulting per-pixel decay field replaces the single global decay value, $d_k$, used in the standard LI calculation (\ref{eq:surface}), allowing spatially varying temporal integration across the image. Finally, the pixel values for a leaky time surface using ER-based decay are defined by
\begin{equation}\label{eq:S^ER}
S_k^{\mathrm{ER}}(x, y) = H_k(x, y) + d_k^{\mathrm{ER}}(x, y) \cdot S_{k-1}^{\mathrm{ER}}(x, y).
\end{equation}

\subsubsection{Adaptive Decay via Laplacian-of-Gaussian}
The Laplacian-of-Gaussian (LoG) operator forms the basis of our second approach through its ability to detect regions of rapid spatial contrast. We first smooth the histogram of new events with a Gaussian kernel to suppress isolated noise spikes, then apply the Laplacian (the sum of second derivatives) to accentuate edges and fine detail. Taking the mean absolute value of the LoG output in patches, we quantify the presence of crisp, well-defined features in different regions. This informs how much we should rely on existing features in the time-surface by modulating the decay rate. 
More concretely, for event window $k$ we build the histogram $H_k$ as in \eqref{eq:hist}. We then convolve $H_k$ with a $3\times3$ LoG kernel, formed from Laplacian kernel $\nabla^2$ and Gaussian kernel $G$ with standard deviation $\sigma=0.25$, to get the edge map

\begin{equation}
E_k(x,y) = \bigl(\nabla^2 G_\sigma * H_k\bigr)(x,y).
\end{equation}

Then, for each patch $P$ of height $h_P$ and width $w_P$, we compute a single sharpness score, $L_P$, by averaging the absolute LoG response over that patch:
\begin{equation}
L_P = \frac{1}{h_P~w_P} \sum_{(x,y)\in P}\bigl|E_k(x,y)\bigr|.
\end{equation}

A patch with a high LoG score indicates sharp incoming features, so the corresponding region on the existing time-surface should be decayed more rapidly. We make the mapping from $L_P$ to decay factor $d_P^{\mathrm{LoG}}(P)\in[0,1]$ via a modified sigmoid:

\begin{equation}
d_P^{\mathrm{LoG}} = \frac{1 + e^{-a~(\tau)}}
                          {1 + e^{-a~(\tau-L_P )}}
\end{equation}

where $\tau$ represents the LoG score where the decay rate is centered, that is, $d_P^{\mathrm{LoG}} \approx 0.5$ at $L_P = \tau$. The parameter $a$ controls the steepness of the function roll-off. Finally, these patch-wise decay values are bilinearly interpolated back to the full image resolution, yielding $d_k^{\mathrm{LoG}}(x,y)$, which again substitutes the global decay, $d_k$, in the LI update formula (\ref{eq:surface}).

\subsubsection{Adaptive Decay via Fast Fourier Transform}
The third adaptive decay method leverages the frequency-domain characteristics of local event activity within each patch by application of the fast Fourier transform (FFT). The intuition is that patches containing sharp edges exhibit greater energy in the high-frequency components of their spatial spectrum. Quantifying the relative contribution of these high-frequency components in patches of the incoming histogram, we apply a localized weighting to its influence on the time-surface by modulating the decay factor.

For each patch, $H_P$, in the histogram of new events, $H_k$, we take the 2D fast Fourier transform, $\mathrm{FFT2}(H_P)$, and compute the power spectrum
\begin{equation}
\mathrm{POW}(H_P) =\bigl|\mathrm{FFT2}(H_{P})\bigr|^2.
\end{equation}

A high-pass filtering of the patch power spectrum is performed by zeroing values within a user defined radius $r\in[0,1]$, normalized against the patch’s largest dimension and centered on the zero-frequency bin. This gives masked power spectrum $\mathrm{POW}_{HF}(H_P)$, whose values are summed to obtain the high-frequency energy. Taking this high-frequency energy as a fraction of the total energy, we define the FFT-based decay factor for each patch
\begin{equation}\label{eqn:d_P^FFT}
d_P^{\mathrm{FFT}} = \frac{\sum \mathrm{POW}_{HF}(H_P)}{\sum \mathrm{POW}(H_P)}.
\end{equation}
These are once again interpolated to a per-pixel decay map $d_{k}^{\mathrm{FFT}}(x,y)$ for each event window $k$, to replace $d_k(x,y)$ in~(\ref{eq:surface}).

\begin{figure*}[t!]
    \centering
    \caption{Sample representations from an event video showing a blink on an otherwise still face. In the FFT examples, grid lines mark patch boundaries and a heatmap generated from the decay value assigned to each patch (before interpolation). With fewer patches, the recursive approach reduces computation while maintaining precise localization of the blink motion, which helps prevent nearby stationary features, such as the eyebrows, from receiving elevated decay values.}
    \label{fig:fft_recurr}
    \setlength{\tabcolsep}{0pt}
    \begin{tabular}{@{}
                    >{\centering\arraybackslash}m{0.3\linewidth}
                    >{\centering\arraybackslash}m{0.3\linewidth}
                    >{\centering\arraybackslash}m{0.3\linewidth}
                    @{}}
    \\[-.75em]
    Histogram & Non-recursive FFT & FFT (Recursive) \\
    \hline
    \multicolumn{3}{>{\centering\arraybackslash}m{0.9\linewidth}}{\includegraphics[width=\linewidth]{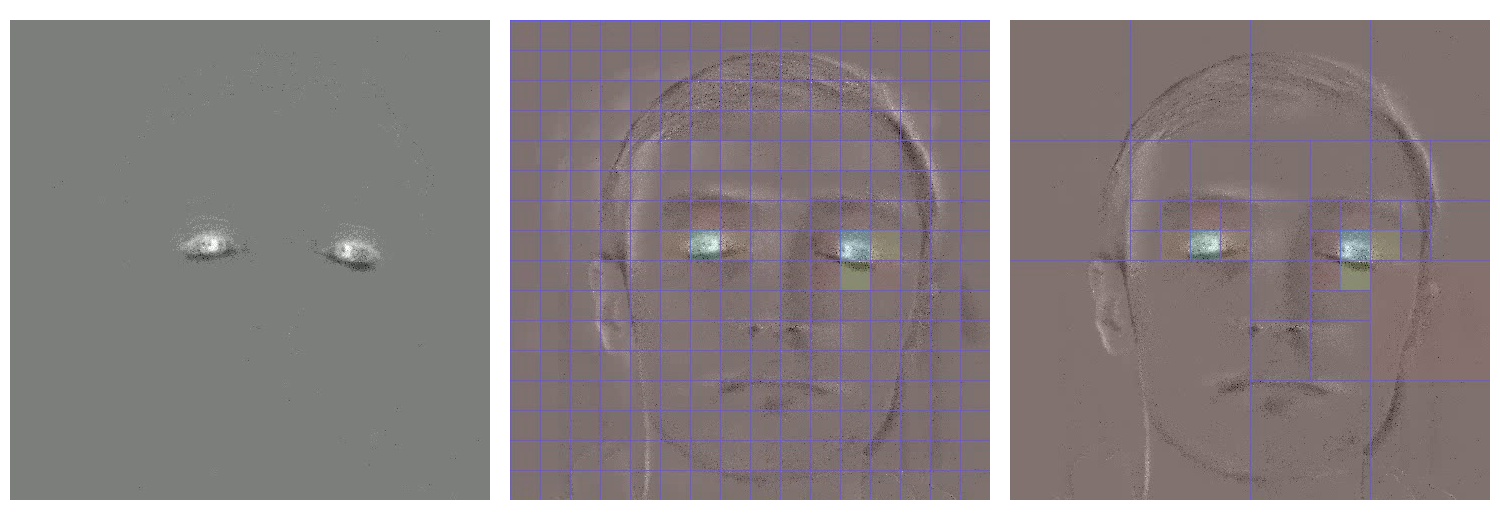} } \\
    \end{tabular}
\end{figure*}

\begin{figure*}[t!]
    \centering
    \caption{Each row shows a different stage of an event video: (a) during rapid head motion, (b) minimal head motion sustained for 0.5 s, and (c) minimal head motion sustained for over 2 s. Columns, left to right, are: histogram representation; time-surface with standard global LI; and time-surfaces generated by the three proposed adaptive integration methods (ER, LoG, FFT).}
    \label{fig:30hz_comp}
    \setlength{\tabcolsep}{0pt}
    \begin{tabular}{@{}>{\centering\arraybackslash}m{0.05\linewidth}
                    @{}>{\centering\arraybackslash}m{0.19\linewidth}
                    @{}>{\centering\arraybackslash}m{0.19\linewidth}
                    @{}>{\centering\arraybackslash}m{0.19\linewidth}
                    @{}>{\centering\arraybackslash}m{0.19\linewidth}
                    @{}>{\centering\arraybackslash}m{0.19\linewidth}
                    @{}}
    \\[-.75em]
        & Histogram & Global LI & LADS-ER & LADS-LoG & LADS-FFT \\ 
    \cline{2-6}
    \\[-.9em]
    (a) & \multicolumn{5}{>{\centering\arraybackslash}m{0.95\linewidth}}{\includegraphics[width=\linewidth]{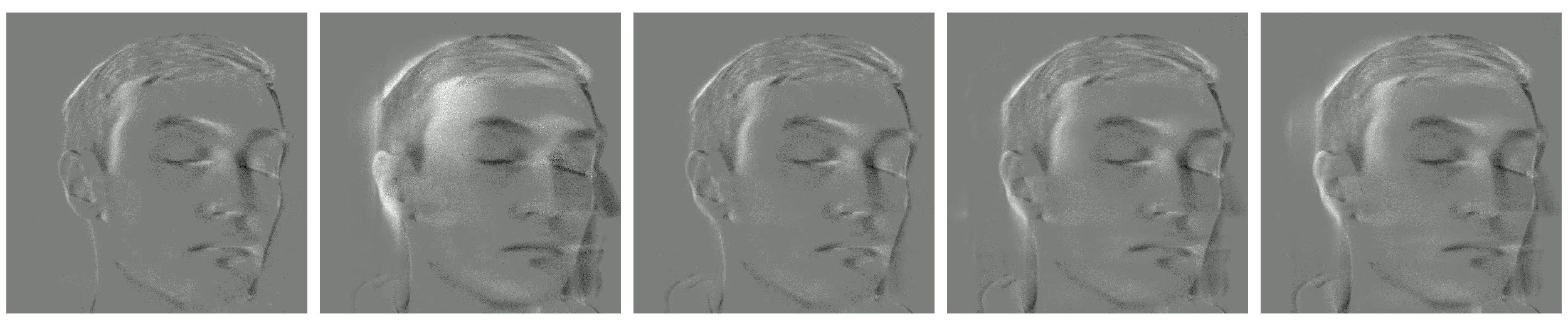} } \\
    (b) & \multicolumn{5}{>{\centering\arraybackslash}m{0.95\linewidth}}{\includegraphics[width=\linewidth]{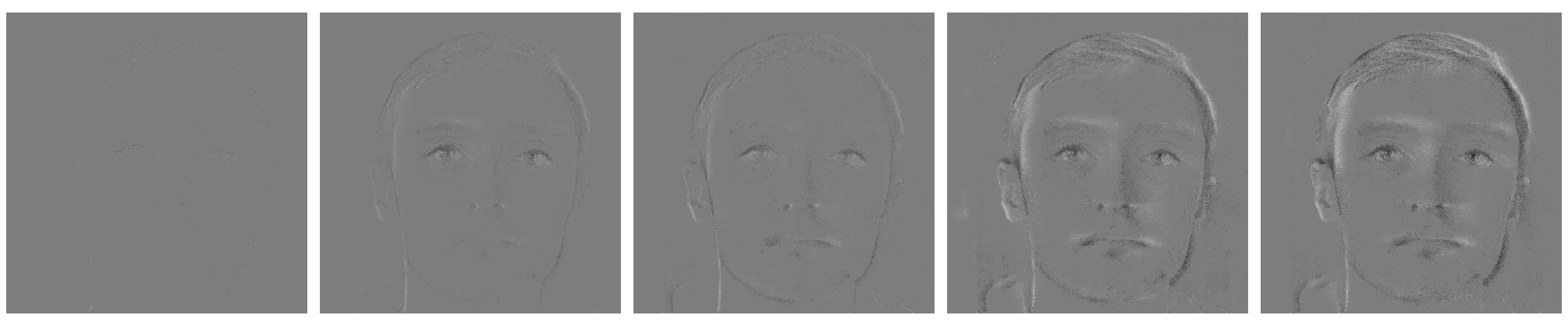} } \\
    (c) & \multicolumn{5}{>{\centering\arraybackslash}m{0.95\linewidth}}{\includegraphics[width=\linewidth]{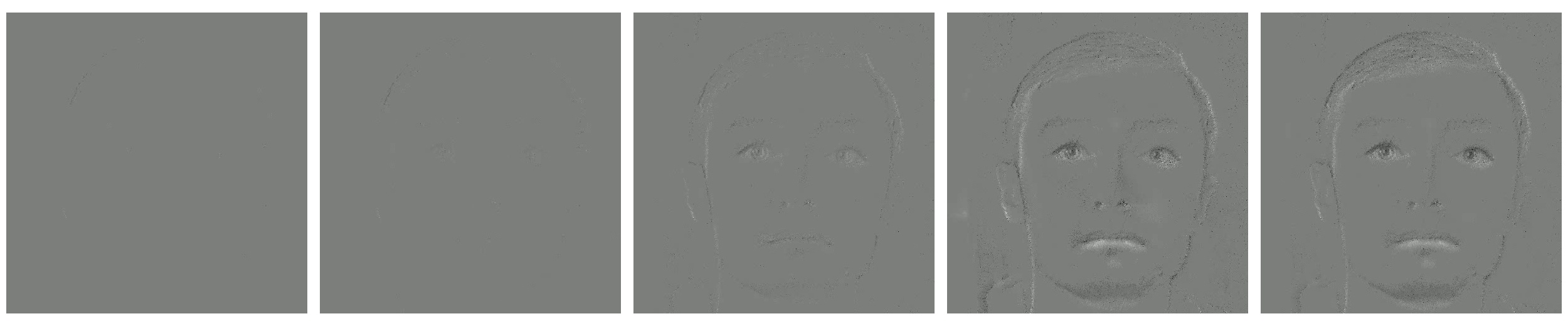} } \\[-.5em]
    \end{tabular}
    
\end{figure*}

Early experiments with this method indicated a strong ability to identify active regions, but also revealed that its computation time was prohibitive for real-time use. Increasing the patch size reduced processing time substantially, though at the cost of a less precise decay map. To address this, a recursive strategy was adopted in which regions are subdivided until the patch score exceeds a threshold $T_d$, or a minimum patch size is reached. This approach, illustrated in Fig. \ref{fig:fft_recurr}, produced suitably localized decay maps while reducing construction time by roughly an order of magnitude. Although validated only through visual inspection, it was adopted as the default implementation of the “FFT” method for the remainder of this work, as otherwise the method would be excluded solely on the basis of computation time. More discussion and data on the timings of the various methods are presented in Section~\ref{sec:results}.

Fig. \ref{fig:30hz_comp} compares the histogram and global-LI representations against the three proposed adaptive methods at three points in the same event video, selected to capture varying scene dynamics. For all tasks and datasets, the patch size was set to one-eighth of the largest image dimension to ensure proportional scaling given the variation in input sizes. For the FFT method, this value defined the minimum patch size used in the recursive subdivision process, preventing further refinement beyond this scale.

\begin{figure*}[b!]
    \centering
    \begin{tabular}{ccc}
     \includegraphics[width=0.3\linewidth]{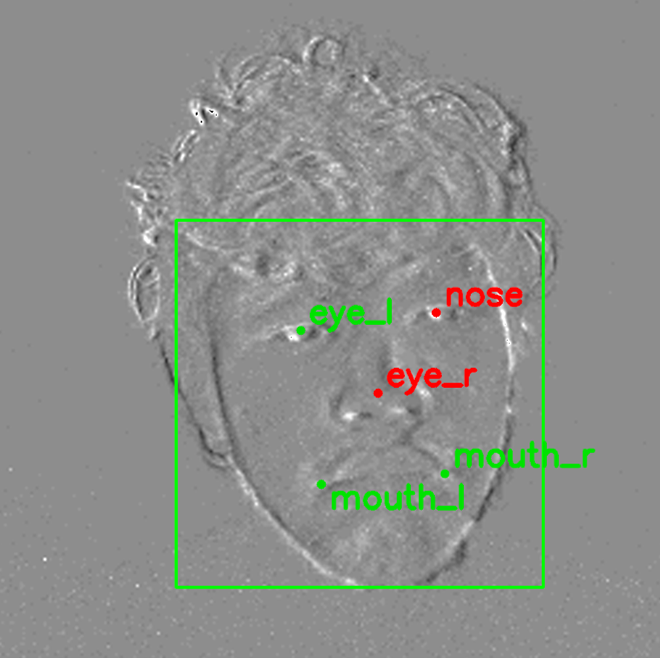} & \includegraphics[width=0.3\linewidth]{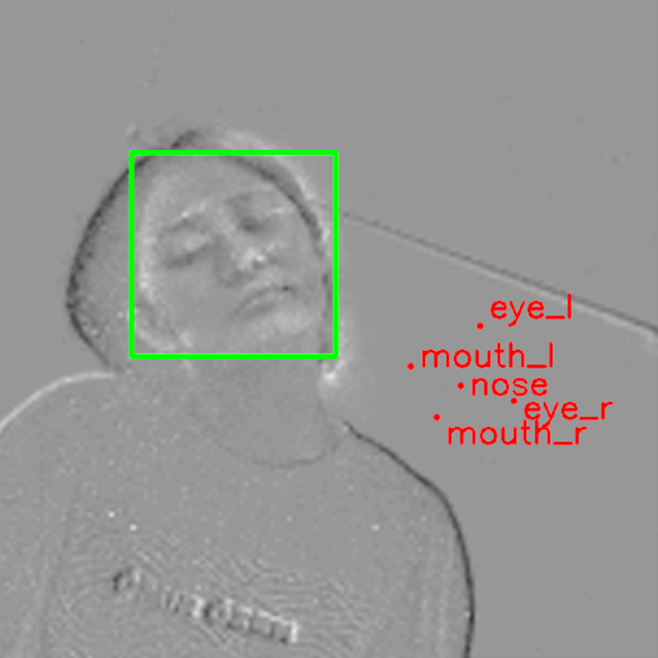} & \includegraphics[width=0.3\linewidth]{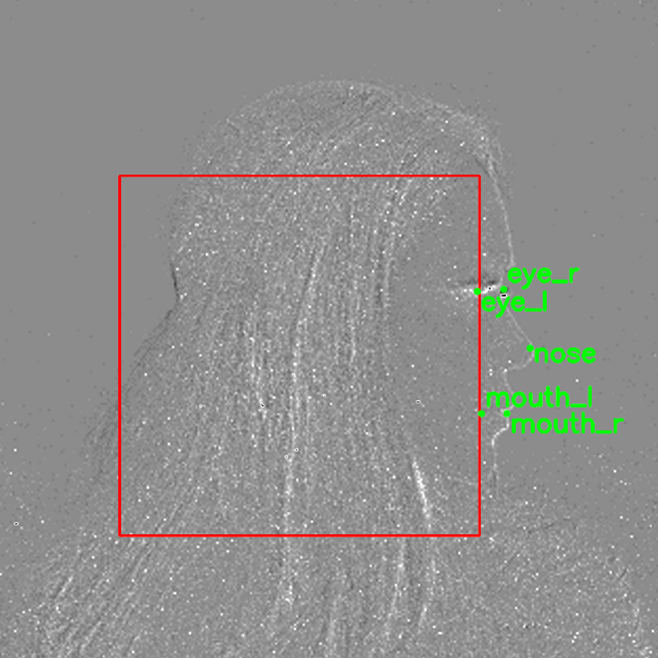} \\
     (a) & (b) & (c) \\
    \end{tabular}
    \caption{Examples of exclusions from the FES dataset with correct elements in green and incorrect elements in red: (a) Inconsistent landmark indexing. (b) Correct bounding box but incorrect landmark positions. (c) Incorrect bounding box but correct landmark positions.}
    \label{fig:fes_mislabels}
\end{figure*}

\subsection{Dataset Description, Cleaning, and Preprocessing}
The FES dataset \cite{Bissarinova2024} is the foundation for all experiments in this work. Recordings were collected using a Prophesee Gen3.1 event-based vision sensor with a spatial resolution of 480×360 and annotated at 30~Hz with a face bounding box and five landmarks (eye centers, nose tip, and mouth corners). Recordings are divided into “lab” and “wild” subsets, where lab samples feature a single participant in a controlled setting, while wild samples include multiple people and unconstrained movement. Only the lab samples were used in this study, as restricting to single-subject recordings greatly simplified the automated cleaning and data loading processes. Lab recordings also constitute the majority of the dataset, and the authors report their results on the lab set independently.

A substantial number of samples were found to contain annotation errors of different forms, such as those with inconsistent landmark indexing and misplaced labels that have been visualized in Fig.~\ref{fig:fes_mislabels}. Further errors included landmarks that froze for several frames before jumping abruptly, multiple contradicting annotations at a single timestamp, and occasionally missing landmarks. 

To address these issues, we developed an automated filtering pipeline combining geometric and temporal consistency checks. Annotations were screened for duplicates and missing landmarks; landmarks outside their bounding box (with a repair step that expanded boxes by up to 10\% of their diagonal); facial topology rules that enforced correct horizontal and vertical ordering of eye and mouth features, with constraints on the nose position based on an inferred head pose; and spatiotemporal consistency, discarding samples where landmark motion diverged from the bounding box motion by more than 20\% of the box diagonal. 

Samples failing any check were excluded, with a small margin of neighboring samples also removed to eliminate residual errors that skirt below thresholds (five samples either side of a failed spatiotemporal consistency check, two for landmarks outside the box). Videos were segmented into continuous clips, and only those at least one second in duration were retained. Finally, clips flagged with high error rates by a pre-trained landmark detector were manually reviewed. The updated set of these revised annotations has been made available online at \href{https://github.com/Paul-Kielty/FES_dataset_revised}{\textcolor{blue}{\underline{https://github.com/Paul-Kielty/FES\_dataset\_revised}}}. The new total duration and sample count are given for each partition of the dataset in Table~\ref{table:dataset}.

\begin{table}[h]
    \centering
    \caption{FES dataset partition details after cleaning}
    \label{table:dataset}
    {\renewcommand{\arraystretch}{2}
    \begin{tabular}{>{\centering\arraybackslash}p{0.12\linewidth}|>{\centering\arraybackslash}p{0.15\linewidth}|>{\centering\arraybackslash}p{0.23\linewidth}|>{\centering\arraybackslash}p{0.15\linewidth}}
        
        \hline
        Partition & \# Subjects & Minutes of Video & \# Samples \\ \hline \hline
        Train & 39 & 283   & 509k \\ 
        Valid & 12 & 79.2  & 143k \\ 
        Test  & 12 & 75.4  & 136k \\ \hline 
        All   & 63 & 437.6 & 788k \\ \hline 
    \end{tabular}
    }
\end{table}

In prior work we combined the FES dataset with others for the detection of facial landmarks and blinks \cite{kielty_lm+blink}. The use of synthetic datasets was in part justified by accumulating network inputs at 30 Hz, which matched the temporal resolution of the source video used for event simulation. In contrast, the present work aims to achieve robust inference at much higher update rates by integrating over much smaller event windows. As synthetic datasets derived from frame-based video may not reliably capture the temporal structure needed for such high-frequency integration, we limit our experiments in this article to real events. This includes the FES dataset but also the Blink dataset used in  \cite{kielty_lm+blink}. Though it contains no landmark annotations, it does provide facial bounding boxes. Moreover, these recordings have a distinct signal profile to FES, due to both the camera characteristics and subject behavior. Visual inspection revealed a much greater prevalence of noise events in FES videos, potentially a result of lower sensor contrast thresholds. Additionally, FES recording has a prescribed task, almost always containing head or camera motions. The Blink dataset, while also consistently focused on faces, makes no guarantee of motion to generate events. These lead to contrasting event rates, measured at approximately 16~event~pixel\textsuperscript{-1}~s\textsuperscript{-1} in the FES test set compared to 1~event~pixel\textsuperscript{-1}~s\textsuperscript{-1} in Blink samples. The Blink dataset was retained for testing only to draw more fair comparisons to other works and to assess method generalization. Some Blink samples were excluded due to missing annotations and gaps in sequences that inhibited the construction of the event surface over time. The final count was 8800 annotated samples across 7 subjects.

\subsection{Network Architecture}

\subsubsection{Facial Landmark Detection}
The landmark detection network architecture in this study is adapted from our previous work \cite{kielty_lm+blink}, where we employed the PIPNet architecture \cite{pipnet}. The network in that article was designed for multi-task prediction (including blink detection) and used a convolutional LSTM module to improve temporal consistency across event sequences. The backbone for feature extraction was MobileNetV2, chosen for its efficiency and demonstrated suitability in event-based settings.

In this study the network is specialized to the FES task: only five landmarks are predicted, and the blink detection head is omitted. To assess the preservation of spatial information by representation choice over a more computationally expensive, recurrent architecture, the previously used convolutional LSTM was also removed. Additionally, the feature extraction backbone is upgraded to MobileNetV3-Large, offering increased representational capacity with minimal computational overhead \cite{mbnetv3}. With these changes, the parameter count of the new landmark detection model was reduced from 9.2M to 3.5M.

\subsubsection{Face Detection}
The face detection network in this study is adapted from the YOLO-based detector of \cite{ryan2021real}. In the original model, a gated recurrent unit was included to improve temporal consistency. As with the landmark model, to allow a clearer assessment of the spatial information preserved by different representations, this recurrent module was removed. During early tests we observed signs of overfitting to the training data, which we attribute to the original detector having been designed to predict both face and eye bounding boxes. Since only face boxes are required in this work, the network was simplified by removing the two largest convolutional layers (layers 10 and 12 as listed in \cite{ryan2021real}). After these changes, the parameter count of the detector fell to 2.6M.

\subsection{Training and Evaluation Protocol}
NME is a widely used metric in facial landmark localization research, as it provides a scale-invariant assessment of prediction accuracy across varying face sizes \cite{pipnet}. In this work, we adopt NME following the convention established by the FES benchmark to ensure comparability with prior studies. For each sample with $L$ landmarks, let $\| \hat{\mathbf{x}}_i - \mathbf{x}_i \| $ denote the Euclidean distance between predicted and ground-truth coordinates, respectively, of the $i$-th landmark. The NME for a single sample is defined as:
\begin{equation}
\mathrm{NME} = \frac{1}{L} \sum_{i=1}^L \frac{ \| \hat{\mathbf{x}}_i - \mathbf{x}_i \| }{ \sqrt{ w_{\mathrm{crop}} \cdot h_{\mathrm{crop}} } }
\end{equation}
where $w_{\mathrm{crop}}$, $h_{\mathrm{crop}}$ are the width and height of the cropped image input, though for our network these will all be square. The final NME is computed as the mean over all test samples. Face detection models are compared by the standard mean average precision (mAP) with an intersection over union threshold of 0.5.

Network inputs are constructed as sequences from consecutive event windows so the time-surfaces have some past events to build from. Otherwise, all representations would be equivalent to histograms. The first 5 event windows from each video were not input to the network, but were still constructed to "warm up" the time-surface. All inputs are cropped to a square region. This is accomplished by expanding the face bounding box to the largest dimension for the landmark detector inputs, and zero-padding where necessary. To improve model robustness and generalization, several data augmentation strategies are applied during training. Each training sequence is randomly mirrored about the vertical axis with a probability of 50\% and random in-plane rotations of up to $\pm$30 degrees. This was where augmentation stopped for the training samples for face detection, but not for landmark detection. They saw random scaling of the bounding box by up to 30\% between sequences, and for individual event representations within a sequence, a smaller translation of up to 3\%, allowing for modest temporal variability. The scaling was followed by random translations with a maximum shift of 10\% for the entire sequence. To prevent the loss of relevant facial information, translations are restricted such that the original face bounding box remains fully visible within the crop.

To mitigate the influence of outlier event counts, all accumulated inputs are clipped to the range $[-5, 5]$ prior to normalization. This prevents single pixels with excessive events from dominating the resulting histogram or time-surface, thereby preserving relative feature visibility across the image.

To allow for faster development, early models were not trained on the full dataset. Instead, a maximum of 75 annotated event windows were taken from each FES video in the training and validation sets. For the training set, these were randomly resampled every epoch. The test sets (FES and Blink) were used in full for all experiments.
Models were trained with each representation, first to run at 30Hz, then 240Hz. The data are annotated at 30Hz, so for higher frequencies, the landmarks and bounding box corners are linearly interpolated. These interpolated landmarks are used in the loss calculation for training and validation, but only the original ground truth labels were used when scoring on the test set. By choosing a higher frequency that is a multiple of 30, we maximize the number of samples with aligned timestamps, and hence maximize the number of testing samples. 

\section{Results and Analysis}
\label{sec:results}

\begin{table}[h]
    \centering
    \caption{Comparison of different event representations for face and landmark detection on FES dataset at 30Hz and 240Hz}
    \label{table:FES_results}
    {\renewcommand{\arraystretch}{2}	
    \begin{tabular}{>{\centering\arraybackslash}m{0.2\linewidth}|
                     >{\centering\arraybackslash}m{0.12\linewidth}
                     >{\centering\arraybackslash}m{0.13\linewidth}|
                     >{\centering\arraybackslash}m{0.12\linewidth}
                     >{\centering\arraybackslash}m{0.13\linewidth}}
        \hline
        \multirow{2}{*}{Representation} & \multicolumn{2}{c|}{Face Detection mAP\textsubscript{50}} & \multicolumn{2}{c}{Landmark NME (\%)}   \\ \cline{2-5}
                                        & 30Hz            & 240Hz                                   & 30Hz          & 240Hz                    \\ \hline \hline
        Histogram                       & 0.921           & 0.829                                   & 2.60          & 4.05                     \\ 
        Global-LI                       & 0.948           & 0.929                                   & 2.37          & 2.76                     \\
        ER                              & 0.955           & 0.942                                   & 2.38          & 2.55                     \\ 
        LoG                             & \textbf{0.957}  & \textbf{0.943}                          & \textbf{2.29} & \textbf{2.52}            \\ 
        FFT                             & 0.950           & 0.932                                   & 2.30          & 2.53                     \\ 
        \hline 
    \end{tabular}}
\end{table}

Results are reported for the Histogram and global-LI as baseline representations, along with the three proposed LADS methods (ER, LoG, FFT). For each of these five representations, the face detection and landmark detection networks were retrained at both 30~Hz and 240~Hz. The twenty resulting models are compared in Table \ref{table:FES_results}. At 30~Hz, the LADS methods all improved on the baseline representations for face detection, with LoG achieving the highest mAP\textsubscript{50} of 0.957 compared to 0.948 for Global-LI. The Histogram representation was notably weaker, with an mAP\textsubscript{50} of 0.921, illustrating the limitations of simple frame accumulation without temporal decay. Landmark error also fell for LoG (2.29\%) and FFT (2.30\%) compared to Global-LI (2.37\%), although ER was marginally worse at 2.38\%. This was the only instance in which an adaptive method did not surpass the baseline, and the difference was just 0.01\%.

Accuracy declined for all methods on FES at 240~Hz, though the reduction was consistently smaller for the adaptive methods across both tasks. The gap between the global-LI baseline and the proposed approaches widened at this higher frequency, indicating that the benefits of local adaptation become more pronounced when fewer events are accumulated per frame. The overall scale of improvement remains moderate in FES, which can be attributed to the consistent head motions performed by participants and the particular sensor bias settings used for these recordings, both of which reduce the need for sustained preservation of features beyond what is provided by the global decay.

The Blink dataset, excluded from training and validation, provides a test of cross-dataset generalization for the ten face detection models. Event rates are far lower in the Blink dataset than in FES, largely due to the camera’s bias settings and greater prevalence of stationary heads. Inputs therefore contain reduced noise but also fewer facial events, making the preservation of past structure essential. This is demonstrated by the increased margin by which the adaptive methods outperformed the others we see in Table \ref{table:Blink_results}. The LoG is the clear winner, maintaining the lowest error across all tests. At 30~Hz it boasts its strongest lead, reaching an mAP\textsubscript{50} of 0.881 compared to 0.778 for ER and 0.741 for FFT. The best of the baselines was the global-LI with 0.552. Though the accuracy is overall lower for all models compared to their corresponding FES scores, the greater gap between the LADS and baseline representations reflects the stronger reliance on adaptive decay when few events define facial features.

\begin{table}[t]
    \centering
    \caption{Comparison of different event representations for face detection on Blink dataset at 30Hz and 240Hz}
    \label{table:Blink_results}
    {\renewcommand{\arraystretch}{2}	
    \begin{tabular}{>{\centering\arraybackslash}m{0.2\linewidth}|
                     >{\centering\arraybackslash}m{0.15\linewidth}
                     >{\centering\arraybackslash}m{0.15\linewidth}}
        \hline
        \multirow{2}{*}{Representation} & \multicolumn{2}{c}{Face Detection mAP\textsubscript{50}}    \\ \cline{2-3}
                                        & 30Hz        &  240Hz         \\ \hline \hline
        Histogram                       & 0.394       &  0.186         \\ 
        Global-LI                       & 0.552       &  0.536         \\
        ER                              & 0.778       &  0.872         \\ 
        LoG                             & \bf{0.881}  & \bf{0.896}     \\ 
        FFT                             & 0.741       &  0.809         \\ 
        \hline
    \end{tabular}}
\end{table}

\newcommand{\imgcell}[1]{
    \raisebox{0pt}[\dimexpr\height+3pt\relax][\dimexpr\depth+3pt\relax]{
        \includegraphics[width=0.925\linewidth]{#1}
    }
}
\begin{figure}[b]
    \centering
    \setlength{\tabcolsep}{0pt}
    \renewcommand{\arraystretch}{1}
    \setlength{\extrarowheight}{0pt}

    \begin{tabular}{@{}
        >{\centering\arraybackslash}m{0.1\linewidth}|
        >{\centering\arraybackslash}m{0.45\linewidth}|
        >{\centering\arraybackslash}m{0.45\linewidth}@{}}
            \hline
            Hz  & FES                                   & Blink                                \\ \hline \hline
            30  & \imgcell{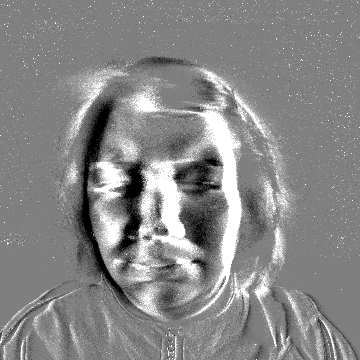}  & \imgcell{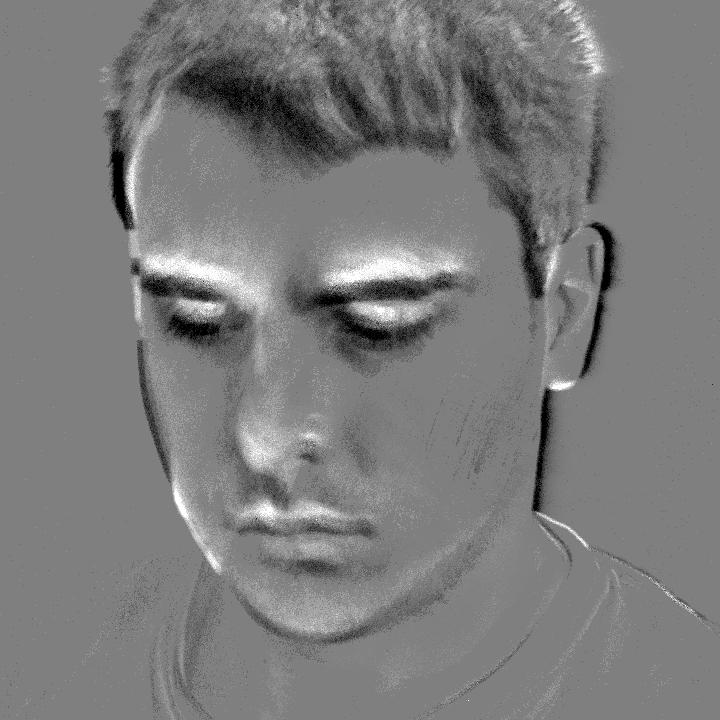}  \\ 
            240 & \imgcell{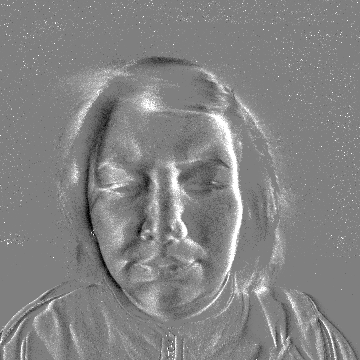} & \imgcell{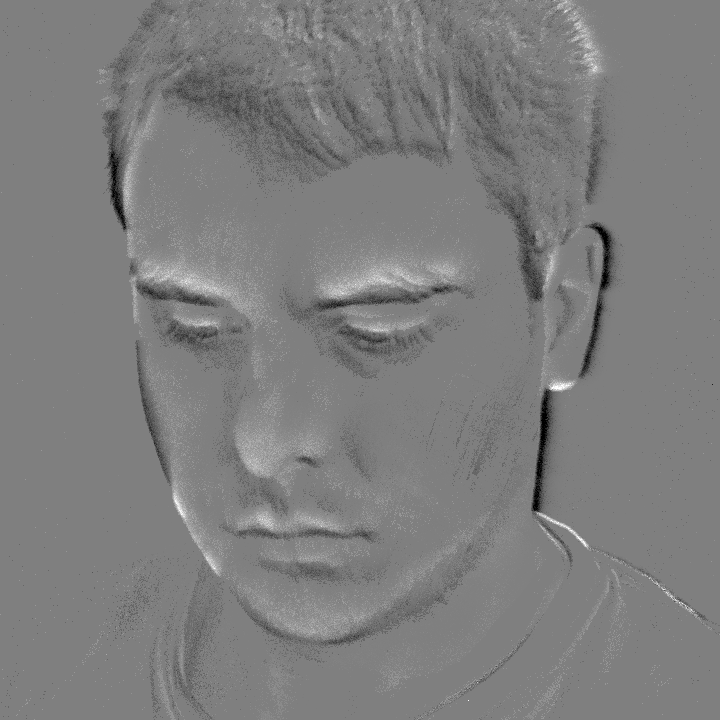} \\ \hline
    \end{tabular}
    \caption{LADS-ER representation of an event window featuring fast head motion from FES and Blink datasets at 30~Hz and 240~Hz.}
    \label{fig:ER_FES_BL}
\end{figure}

Testing the Blink dataset at 240~Hz, the adaptive methods improved slightly compared to their own 30~Hz results, with LoG rising from 0.881 to 0.896. This contrasts with the drop seen in FES and likely stems from differences in how the Blink dataset aligns with the FES training data at each frequency. First, increasing the frequency in FES slices head movements into smaller temporal segments, producing consecutive changes that are closer in scale to those in Blink, where motion is minimal. A second factor lies in how decay is applied. Each new window of events is added directly to the surface, with decay applied only to the existing state. At 30~Hz in FES, the large number of events per window produces more heavily cluttered inputs, a problem visible in Fig.~\ref{fig:ER_FES_BL} where the 30~Hz FES frame is obscured by over-accumulation of motion events. At 240~Hz, these same motions are broken into smaller batches that can be decayed more regularly. For the standard representations, this means a broadly comparable reduction in event density across both datasets, since older events are discarded more frequently regardless of activity level. In contrast, LADS adapts its decay rate to the local dynamics: in FES it typically decays more aggressively to manage the dense motion, while in Blink, where far fewer events occur, it preserves more of the available activity. Consequently, training on 30~Hz FES requires fitting to heavily saturated inputs that differ markedly from Blink, whereas at 240~Hz the FES representations are closer in scale and profile to those from Blink. This narrows the cross-dataset gap and explains why the 240~Hz models generalize better to the Blink test set, even though accuracy within FES itself is higher at 30~Hz.

These findings do not imply that higher frequency is inherently better for face detection. In our study, the effect arises because higher-frequency processing with LADS brings the FES representations closer to those of Blink, narrowing the gap caused by their very different event densities. More generally, in environments with variable motion and event rates, such convergence of input profiles by higher-frequency processing with LADS may offer a practical advantage for cross-domain generalization.

Table~\ref{table:timings} reports the mean construction times for each representation at 30 Hz and 240 Hz. Histogram and Global-LI were fastest, each requiring well under 1 ms per integration step. The adaptive methods incurred a measurable overhead, with recursive FFT the most expensive at 1.86 ms for 30 Hz. By contrast, the non-recursive FFT exceeded 10 ms per step at both frequencies, showing little improvement with higher frequency since its subdivision is applied uniformly. The recursive strategy adapts more intelligently, reducing subdivision when fewer events fall within a window, and thus achieving a speedup at 240 Hz. Overall, while the adaptive methods require noticeably more time than the baselines, their costs remain compatible with real-time deployment.        

Among the adaptive variants, LoG emerged as the strongest performer overall, combining high accuracy with competitive speed. Its main limitation is in tuning; since the LoG score is effectively unconstrained and must be mapped to decay values via an arbitrarily chosen falloff; the process of parameter selection is less straightforward. FFT offers a different balance, showing greater robustness than ER in noisy data and benefiting from parameters that are tightly bounded between 0 and 1; making it easier to converge on suitable values. However, FFT was significantly slower than the other methods and performed less effectively in sparse conditions. ER provides a useful middle ground, offering fast computation and relatively accessible tuning. Its decay value follows the same form as in leaky integration, with the addition of a parameter that can be set directly from measurements of the event rate. At the same time, its simplicity leaves it highly sensitive to noise and continuously misfiring "hot pixels"~\cite{hot_pixels}.

\begin{table}[h]
    \centering
    \caption{Mean time to construct the different event representations for each frame of the video shown in Fig. \ref{fig:30hz_comp} at 30Hz and 240Hz. System hardware: NVIDIA GeForce RTX 2080 Ti, Intel Core i9-9960X @ 3.10~GHz, 128~GiB DDR4 @ 3~GHz}
    \label{table:timings}
    {\renewcommand{\arraystretch}{2}	
    \begin{tabular}{>{\centering\arraybackslash}m{0.32\linewidth}|
                     >{\centering\arraybackslash}m{0.15\linewidth}
                     >{\centering\arraybackslash}m{0.15\linewidth}}
        \hline
        \multirow{2}{*}{Representation} & \multicolumn{2}{c}{Time (ms)}    \\ \cline{2-3}
                                        & 30Hz      & 240Hz      \\ \hline \hline
        Histogram                       & \bf{0.50} & \bf{0.24}  \\ 
        Global-LI                       & 0.59      & 0.28       \\
        ER                              & 1.19      & 0.60       \\ 
        LoG                             & 0.95      & 0.52       \\ 
        FFT                             & 1.86      & 0.61       \\ 
        FFT (non-recursive)             & 11.15     & 10.51      \\ 
        \hline
    \end{tabular}}
\end{table}

\begin{table}[!b]
    \centering
    \caption{Parameter counts and performance of event based face detectors (by mAP\textsubscript{50}) and landmark detectors (by NME) of increasing inference frequencies, trained and tested on FES data.}
    \label{table:freq_comparison}
    \setlength{\tabcolsep}{0pt}
    {\renewcommand{\arraystretch}{2}	
    \begin{tabular}{
                    >{\centering\arraybackslash}m{0.16\linewidth}|
                    >{\centering\arraybackslash}m{0.1\linewidth}|
                    >{\centering\arraybackslash}m{0.25\linewidth}|
                    >{\centering\arraybackslash}m{0.15 \linewidth}|
                    >{\centering\arraybackslash}m{0.16\linewidth}|
                    >{\centering\arraybackslash}m{0.16\linewidth}}
    
        \hline
        Frequency    & Model & Event Input & \# Params (M) & Face mAP\textsubscript{50} & Landmark NME (\%) \\ \hline \hline
        
        \multirow{2}{*}{20 Hz}  & \cite{Bissarinova2024} & Histogram    & 24.1    & 0.973   & 2.44 \\
                                & \cite{spiking_face}    & Event Spikes & 1.2     & 0.581   & -    \\ \hline \hline
        \multirow{5}{*}{30 Hz}  & \cite{Bissarinova2024} & Histogram    & 24.1    & 0.370   & 6.35 \\ 
                                & \cite{spiking_face}    & Event Spikes & 1.2     & 0.557   & -    \\
                                & \cite{kielty_lm+blink} & Global-LI    & 9.2     & -       & 4.68 \\
                                & Ours                   & LADS-LoG     & 2.6     & 0.969   & -    \\
                                & Ours                   & LADS-LoG     & 3.5     & -       & 2.21 \\ \hline \hline
        \multirow{2}{*}{240 Hz} & Ours                   & LADS-LoG     & 2.6     & 0.966   & -    \\
                                & Ours                   & LADS-LoG     & 3.5     & -       & 2.44 \\ \hline
    \end{tabular}
    }    
\end{table}

To assess the full potential of the proposed approach, we selected LoG as the best-performing variant (offering the strongest balance of accuracy and speed once tuned) and trained new LADS models at both 30~Hz and 240~Hz on the complete FES training and validation sets, rather than the earlier limit of 75 event windows per video that had been used to accelerate development. Table~\ref{table:freq_comparison} compares the performance and parameter counts of these models and other face and landmark detectors trained on FES. The 20 Hz histogram model of \cite{Bissarinova2024} demonstrated strong results in face detection (0.973~mAP\textsubscript{50}) and landmark detection (2.44\% NME), but required a large 24.1~M parameter network. At 30~Hz, however, the same representation fell off sharply to 0.370~mAP\textsubscript{50} and 6.35\% NME, highlighting how quickly the histogram degrades for FES once the prediction frequency rises above the relatively long 20~Hz integration windows. The Global-LI model from our own previous work \cite{kielty_lm+blink} provided a smaller and more robust alternative, reducing network size to 9.2~M parameters while improving landmark accuracy to 4.68\%. A separate line of work we explored was the use of hybrid spiking networks that bypass the usual representation step entirely \cite{spiking_face}. While remarkably efficient, with only 1.2~M parameters for face detection, these models suffered in accuracy. Especially at higher temporal resolutions, with the mAP\textsubscript{50} falling sharply to 0.09 when events were input in 5~ms duration windows (200~Hz).

By contrast, our LADS-LoG models achieved state-of-the-art landmark accuracy at 30~Hz with an NME of 2.21\% and a network of only 3.5~M parameters. The representation was similarly impressive in face detection, achieving an mAP\textsubscript{50} of 0.969 at 30~Hz. This is only 0.004 below the best FES face detection result, which was achieved by the 20~Hz histogram model of \cite{Bissarinova2024} using nearly ten times as many parameters. Importantly, these lightweight networks maintained competitive accuracy even when operating at 240~Hz, with 2.44\% NME and 0.966~mAP\textsubscript{50}. These results establish new benchmarks for both face detection and landmark localization in events at higher frequencies, demonstrating that adaptive integration not only improves accuracy but also enables efficient high-rate prediction with substantially reduced model size.

\section{Conclusion}
This work introduced LADS, a set of event-based representations that adapt temporal decay locally according to scene dynamics. Evaluations on face detection and facial landmark detection demonstrated that LADS improves performance over both global-LI and histogram baselines across a range of event densities, with the most pronounced gains occurring in sparse-event conditions where maintaining facial structure is challenging.

The proposed approaches maintain accuracy far better than global-decay methods at higher update rates, where fewer events are accumulated per frame. By preserving relevant structure under these conditions, they enable the use of high-frequency processing without the steep performance penalties observed in standard representations, allowing better exploitation of the event camera’s high temporal resolution. This makes it possible to capture fast or transient motions more reliably, while also supporting significant reductions in network complexity. By retaining features in the input representation itself, LADS reduces the need for heavy recurrent components or other temporal-memory mechanisms, making it possible to deploy lighter architectures without sacrificing accuracy.

The combination of higher-frequency capability, accuracy gains, and reduced architectural demands makes LADS a practical choice for event-based driver monitoring and other in-cabin sensing tasks. More broadly, adapting event integration according to local signal characteristics offers a path toward event-based vision systems that are both efficient and robust across the diverse conditions encountered in real-world~use. 

\subsection{Limitations and Challenges}
While LADS improves performance across a variety of event densities and enables effective high-frequency processing, several limitations remain. The evaluations were conducted on datasets with specific motion profiles, sensor configurations, and parameter settings. The FES dataset features consistent head motions and high event rates, while the Blink dataset has very low event rates due to its bias settings and minimal motion, and each was processed using different decay parameters. Real-world conditions often fall between these extremes or vary significantly over time, which may affect the relative advantage of adaptive decay. Broader testing across a wider range of datasets, event camera models/bias profiles, and conditions will be necessary to fully assess robustness.

The present work examined only two tasks. Expanding evaluation to a broader set of event-based vision problems with differing spatial and temporal dependencies would help to better characterize the versatility and limitations of LADS.

A further limitation is that LADS requires manual setting of the decay parameters that control the degree of temporal preservation. Although an effort was made to ensure fair comparisons in our experiments, parameter selection was guided only by visual inspection of sample outputs. This approach, while practical, may not yield optimal values across all conditions, and more systematic or automated tuning could further improve performance.

\subsection{Future Work}
Further work will explore how LADS performs in a wider spectrum of event-based vision problems and recording conditions, particularly those that differ substantially from the datasets used here. Automating the selection of decay parameters, or allowing them to adapt dynamically during inference, could improve performance without manual intervention. In parallel, investigating more efficient software and hardware implementations will help extend LADS’s suitability to resource-limited platforms.

\section{Acknowledgment}
This research was conducted with the financial support of the Irish Research Council under Grant Agreement No. IRCLA/2023/1992, the Research Ireland Centre for Research Training in Digitally-Enhanced Reality (d-real) under Grant No. 18/CRT/6224, and Science Foundation Ireland under Grant Agreement No. 13/RC/2106\_P2 at the ADAPT SFI Research Centre at the University of Galway. All of the data used in this paper were collected with informed consent and in compliance with ethical guidelines.

\appendices
\section{Implementation Details} 
Some additional details on the LADS implementation:
\begin{itemize}
    \item Overlapping patches could potentially provide more localized and smoother decay values, but considering the processing speed to be a more limiting factor in this work, the associated increase in computational effort was not justified. 
    \item If total energy, the denominator in equation (\ref{eqn:d_P^FFT}) is 0, the decay factor $d_{P}^{FFT}$ is taken as 1. 
    \item Also regarding LADS-FFT, for the zero-frequency to be centered in the power spectrum (required to apply the circular zero masking as required), an FFT-shift operation is required. Instead of doing this for each patch, we pre-compute a shifted mask for each possible patch size.
\end{itemize}

The parameters used for the representation of each dataset are given in Table \ref{table:LADS_params}.
\begin{table}[H]
    \centering
    \caption{Parameters for all decay strategies and datasets at 30Hz and 240Hz.}
    \label{table:LADS_params}
    \renewcommand{\arraystretch}{2.5}
    \begin{tabular}{|>{\centering\arraybackslash}m{0.08\linewidth}|
                     >{\centering\arraybackslash}m{0.14\linewidth}|
                     >{\centering\arraybackslash}m{0.05\linewidth}
                     >{\centering\arraybackslash}m{0.05\linewidth}|
                     >{\centering\arraybackslash}m{0.05\linewidth}
                     >{\centering\arraybackslash}m{0.05\linewidth}|
                     >{\centering\arraybackslash}m{0.05\linewidth}
                     >{\centering\arraybackslash}m{0.05\linewidth}|}
        \cline{2-8}
        \multicolumn{1}{c|}{} & Global-LI  & \multicolumn{2}{c|}{ER} & \multicolumn{2}{c|}{LoG} & \multicolumn{2}{c|}{FFT} \\ \cline{2-8}
        \multicolumn{1}{c|}{} & $\tau$ & $\tau$  & $\lambda_{0}$ & $\tau$ & $a$              & $r$   & $T_{d}$      \\ \hline
        FES 30Hz              & 0.05   & 0.05    & 16            & 12.5   & 0.25             & 0.25  & 0.5            \\ 
        FES 240Hz             & 0.05   & 0.05    & 16            & 12.5   & 0.25             & 0.05  & 0.5            \\ 

        Blink 30Hz            & 0.2    & 0.2     & 2             & 7.5    & 0.75             & 0.01  & 0.9            \\ 
        Blink 240Hz           & 0.2    & 0.2     & 2             & 7.5    & 0.75             & 0.01  & 0.9            \\ \hline
        
    \end{tabular}
\end{table}

\section{FES Dataset Filtering}
For each FES video, the bounding box and landmark annotations at each timestamp were assessed using the following criteria:
\begin{itemize}
    \item \textbf{Annotation count per-timestamp:} Some samples were found to have multiple label entries for a single timestamp (when only one subject is present in the video).
    \item \textbf{Landmark count:} Some samples with fewer than 5 landmark positions were detected.
    \item \textbf{Landmarks within bounding box:} The landmark positions were checked to ensure they are within the bounding box. However, because correct landmarks were sometimes excluded due to a bounding box that was slightly too small, an additional repair check was implemented: if landmarks were outside the bounding box by less than 10\% of the diagonal length, the box dimensions were extended to exactly include the landmarks. Otherwise, the sample was discarded.
    \item \textbf{Facial topology:} To detect instances of inconsistent landmark indexing, a set of rules were created to validate the structure of the face. First, ensure that the left mouth and eye landmarks are actually positioned to the left of the corresponding right landmark. A similar check compared the vertical positions of each eye landmark to the mouth landmark on the same side. Verification of the nose position first required gathering some information of the head pose by comparing vertical and horizontal distances between facial feature pairs. When the is primarily turned to one side (yaw), determined by the vertical distances dominating the horizontal, the nose landmark is required to have a vertical position between the average eye level and average mouth level. When the head is primarily facing up or down (pitch), detected by dominant horizontal differences, the horizontal component of nose landmark must be fall between the left-side and right-side facial landmarks.
    \item \textbf{Spatiotemporal consistency:} This check ensures that facial landmarks and bounding boxes move together cohesively to catch annotation errors where landmarks are frozen or lag behind face movement. Inconsistencies between landmark and bounding box movements are detected across consecutive labels by computing displacement vectors for both elements. Samples are flagged where the absolute difference between the landmark displacement and bounding box displacement exceeds 20\% of the bounding box diagonal.
\end{itemize}


\bibliographystyle{ieeetr}
\bibliography{main}

\begin{IEEEbiography}[{\includegraphics[width=1in,height=1.25in,clip,keepaspectratio]{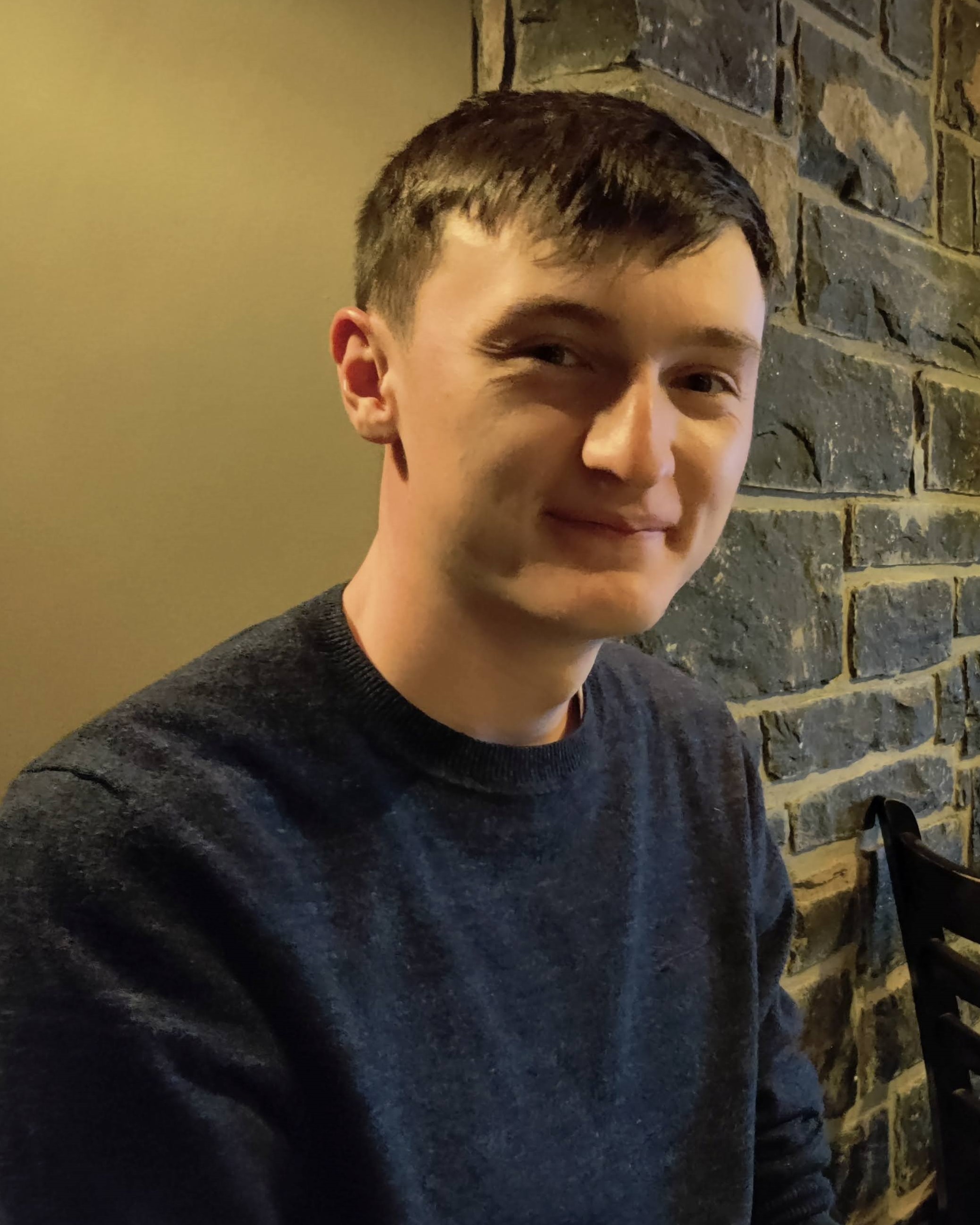}}]{PAUL KIELTY} received the B.E. in electronic and computer engineering from the University of Galway in 2021. He is now pursuing the Ph.D degree with the University of Galway and the ADAPT SFI Research Centre. His research is focused on deep learning methods with neuromorphic vision, with particular interest in facial analytics and driver monitoring tasks.
\end{IEEEbiography}

\begin{IEEEbiography}[{\includegraphics[width=1in,height=1.25in,clip,keepaspectratio]{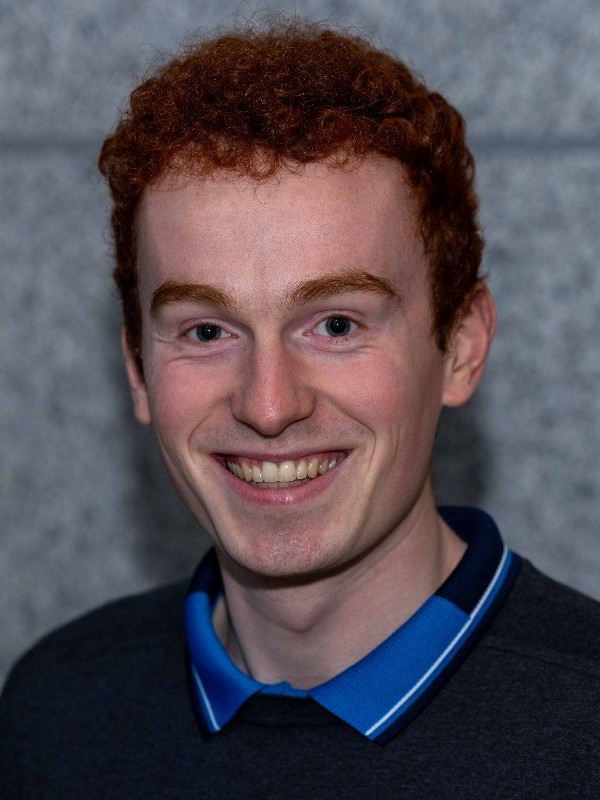}}]{Timothy Hanley} (Graduate Student Member, IEEE) received the B.E. degree in Electronic and Computer Engineering from the University of Galway in 2023, where he is currently undertaking the Ph.D degree in Computer Science with the D-Real SFI Research Centre. His research is focused on neuromorphic vision and deep learning, with particular interest in driver monitoring and non-contact sensing applications. 
\end{IEEEbiography}

\begin{IEEEbiography}[{\includegraphics[width=1in,height=1.25in,clip,keepaspectratio]{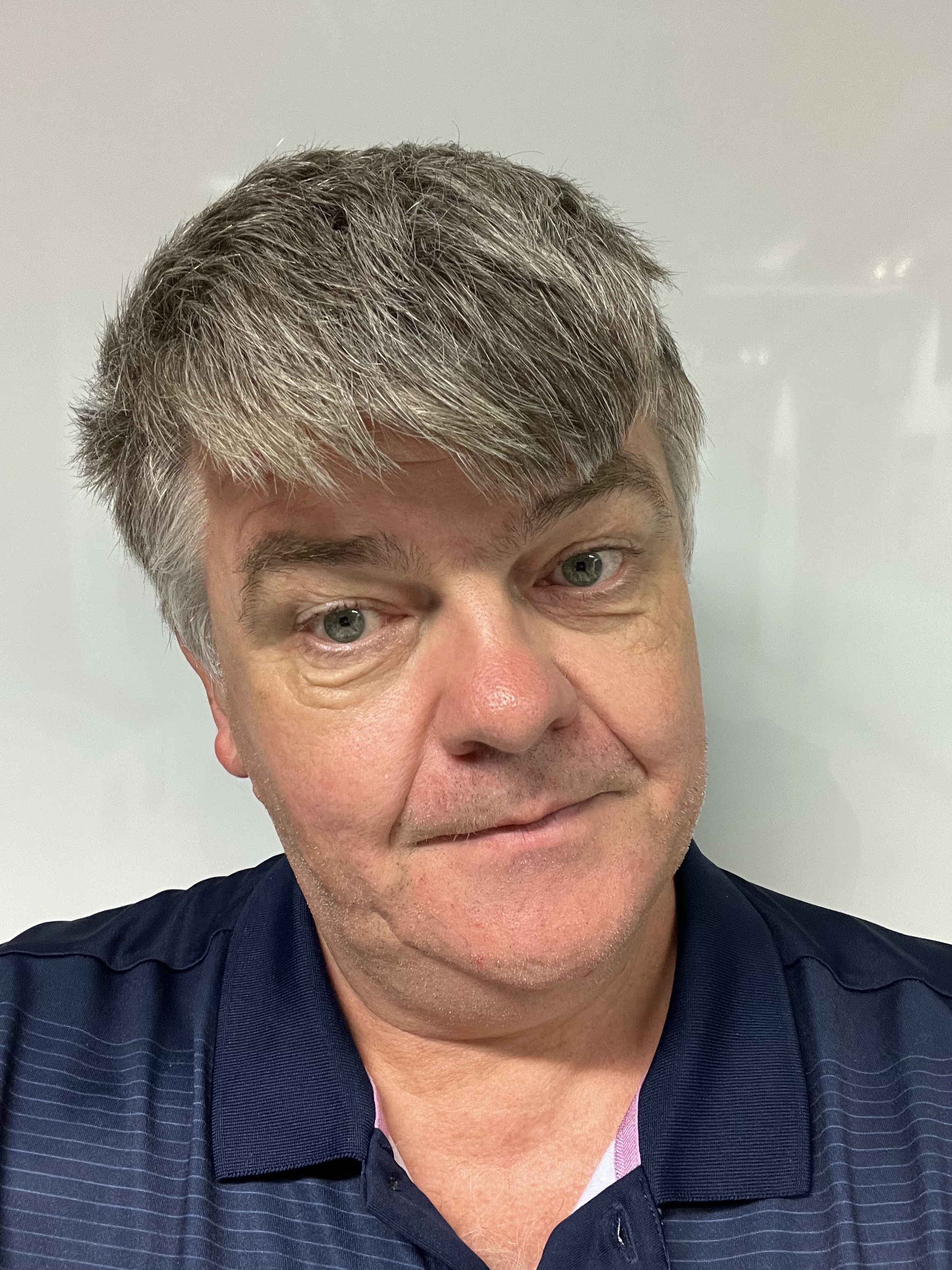}}]{PETER CORCORAN} (Fellow, IEEE) holds the Personal Chair in electronic engineering at the College of Science and Engineering, University of Galway. He is currently an IEEE Fellow recognized for his contributions to digital camera technologies, notably in-camera red eye correction and facial detection. He was a co-founder of several start-up companies, notably FotoNation Ltd. He has over 600 technical publications and patents, over 100 peer-reviewed journal articles, 120 international conference papers, and a co-inventor of more than 300 granted U.S. patents. He has been a member of the IEEE Consumer Electronics Society for over 25 years. He is the Editor-in-Chief and the Founding Editor of IEEE Consumer Electronics Magazine.
\end{IEEEbiography}

\end{document}